\documentclass{article}

\usepackage{arxiv}

\usepackage[numbers, compress]{natbib}
\usepackage[utf8]{inputenc} 
\usepackage[T1]{fontenc}    
\usepackage{hyperref}       
\usepackage{url}            
\usepackage{booktabs}       
\usepackage{amsfonts}       
\usepackage{nicefrac}       
\usepackage{microtype}      
\usepackage{xcolor}         

\usepackage{amsmath}
\usepackage{amssymb}
\usepackage{amsfonts}
\usepackage{amsthm}
\usepackage{mathrsfs}
\usepackage{graphicx}
\usepackage{layouts}
\usepackage[amsstyle]{cases}
\usepackage{nicefrac}
\usepackage{tabularx}
\usepackage{subcaption}
\usepackage[noend]{algpseudocode}
\usepackage{algorithm}
\usepackage[shortlabels]{enumitem}
\setlist{nosep,leftmargin=*}

\usepackage[colorinlistoftodos]{todonotes}

\definecolor{linkblue}{RGB}{50,100,170}
\usepackage{hyperref}
\hypersetup{
    pdfborder={0 0 0},
    colorlinks=true,
    urlcolor=linkblue,
    linkcolor=linkblue,
    citecolor=linkblue,
    bookmarks=true,
    bookmarksnumbered=true
}

\usepackage{tikz}
\usetikzlibrary{decorations.pathreplacing, positioning, shapes, arrows.meta, intersections}
\tikzset{>={Latex[length=1.5mm, width=1mm]}}
\usepackage{tikzsymbols}

\definecolor{d61green}{RGB}{48,184,136}
\definecolor{d61blue}{RGB}{0,169,206}
\definecolor{d61plum}{RGB}{109, 32, 119}
\definecolor{d61oceanblue}{RGB}{0, 75, 135}
\definecolor{myred}{RGB}{200, 100, 100}

\newcommand{\x}{\boldsymbol{x}}
\newcommand{\z}{\boldsymbol{z}}

\newcommand{\W}{\boldsymbol{W}}

\newcommand{\y}{\boldsymbol{y}}

\renewcommand{\b}{\boldsymbol{b}}

\newcommand{\g}{\boldsymbol{g}}

\newcommand{\thetabf}{\boldsymbol{\theta}}
\newcommand{\phibf}{\boldsymbol{\phi}}

\title{Bayesian Neural Network Inference via Implicit Models and the Posterior Predictive Distribution}
\date{}
\author{%
    Joel Janek Dabrowski \\
    Data61, CSIRO, \\
    Australia \\
    \texttt{Joel.Dabrowski@data61.csiro.au}
    \And
    Daniel Edward Pagendam \\
    Data61, CSIRO, \\
    Australia \\
    \texttt{Dan.Pagendam@data61.csiro.au}
}

\begin{document}
    
    \maketitle
    
    \begin{abstract}
        We propose a novel approach to perform approximate Bayesian inference in complex models such as Bayesian neural networks. 
        The approach is more scalable to large data than Markov Chain Monte Carlo, it embraces more expressive models than Variational Inference, and it does not rely on adversarial training (or density ratio estimation).
        We adopt the recent approach of constructing two models: (1) a primary model, tasked with performing regression or classification; and (2) a secondary, expressive (e.g. implicit) model that defines an approximate posterior distribution over the parameters of the primary model.  
        However, we optimise the parameters of the posterior model via gradient descent according to a Monte Carlo estimate of the posterior predictive distribution -- which is our only approximation (other than the posterior model).  
        Only a likelihood needs to be specified, which can take various forms such as loss functions and synthetic likelihoods, thus providing a form of a likelihood-free approach. 
        Furthermore, we formulate the approach such that the posterior samples can either be independent of, or conditionally dependent upon the inputs to the primary model. 
        The latter approach is shown to be capable of increasing the apparent complexity of the primary model. 
        We see this being useful in applications such as surrogate and physics-based models.
        To promote how the Bayesian paradigm offers more than just uncertainty quantification, we demonstrate: uncertainty quantification, multi-modality, as well as an application with a recent deep forecasting neural network architecture.
    \end{abstract}
    
    

\section{Introduction}
\label{sec:introduction}

The Bayesian statistical paradigm offers a principled approach to representing uncertainty in both a model and its predictions by treating the model parameters as latent random variables.  Inference of the latent parameters is, however, generally intractable in complex and nonlinear models such as Bayesian Neural Networks (BNNs).  Approximate inference approaches are employed, with Variational Inference (VI) and Markov Chain Monte Carlo (MCMC) being the most popular \cite{jospin2022hands,abdar2021review}. The challenges with these approaches involve balancing complexity and expressiveness.

MCMC methods, such as Hamiltonian Monte Carlo (HMC) \cite{neal2011mcmc}, provide accurate samples from the posterior and are considered the gold standard approach for inference \cite{goan2020bayesian}. This however comes at the cost of computational complexity and poor scalability due to burn-in periods, sample rejection, and the requirement to process an entire dataset in order to make a new proposal. Stochastic gradient MCMC methods (e.g. \cite{welling2011bayesian,chen2014stochastic}) address this by rather performing MCMC through ``noisy backpropagation''. Such approaches have however been shown to compromise the scalability of HMC \cite{betancourt2015fundamental}.

VI approaches rely on optimisation to provide a tractable and scalable approach. They have enjoyed extensive application in deep learning models such as the Variational Auto-Encoder (VAE) \cite{kingma2014autoencoding} and the BNN \cite{blundell2015Weight}. However, the accuracy of VI is compromised when the variational family is restricted to simple parametric distributions as to promote tractability. These parametric distributions generally lack the expressiveness required for BNNs. Furthermore, VI can tend to underestimate the variance \cite{blei2017variational,bishop2006pattern}; especially under the commonly adopted Mean Field VI (MFVI) assumption.

Recently, several studies have considered replacing parametric distributions with highly expressive implicit distributions \cite{krueger2017bayesian,pawlowski2017implicit,wang2018adversarial,henning2018approximating}. Implicit distributions include the generators in Generative Adversarial Networks \cite{goodfellow2014generative} and have relation to Normalising Flows \cite{papamakarios2021normalizing}. Direct application of VI to implicit distributions is however challenging as these distributions cannot be evaluated, but only sampled from. To overcome this, the problem is reframed into an adversarial problem where a density ratio within the Evidence Lower Bound (ELBO) is approximated with a discriminator \cite{huszar2017variational}. The discriminator is only equal to the density ratio if it is a Bayes-optimal classifier, thus requiring a highly expressive discriminator and increased computational resources \cite{yin2018semi}. Furthermore, density ratio estimation is also challenging in high-dimensional spaces \cite{sugiyama2012density}. 

To avoid density ratio estimation, alternative approaches have been proposed by defining semi-implicit variational approximations \cite{yin2018semi,titsias2019unbiased,moens2021efficient}. These approaches use a hierarchical framework where the variational distribution is explicit, but an implicit mixing distribution is imposed on the parameters of the variational distribution. The approach however does not directly optimise using the ELBO, but a surrogate thereof \cite{yin2018semi,titsias2019unbiased}. To directly optimise according to the ELBO, \citet{titsias2019unbiased} use and MCMC approach, which is however computationally expensive.

The commonality between the discussed approaches is that they all strive to conform to the MCMC or VI approach. We aim to devise an alternative which is both expressive and scalable to large data.

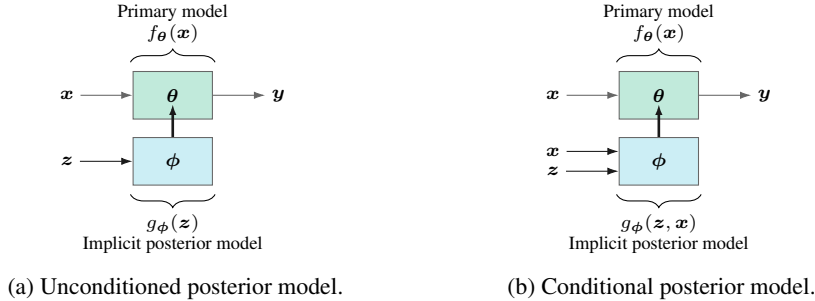
\begin{figure}[!tb]
    \centering
    \begin{subfigure}{0.45\columnwidth}
        \centering

\def\horisep{40pt}
\def\vertsep{25pt}
\def\horisepb{12pt}
\def\vertsepb{12pt}

\begin{tikzpicture}
    \begin{scriptsize}
        
        \tikzstyle{every path}=[->]
        
        \node[] (x) at (-1*\horisep, 0*\vertsep) {$\x$};
        \node[] (y) at (1*\horisep, 0*\vertsep) {$\y$};
        \node[] (z) at (-1*\horisep, -1*\vertsep) {$\z$};
        \node[rectangle, fill=d61green!30, draw=black!60, minimum width=30pt, minimum height=18pt] (n1) at (0*\horisep, 0*\vertsep+0pt) {$\thetabf$};
        \node[rectangle, fill=d61blue!20, draw=black!60, minimum width=30pt, minimum height=18pt] (n2) at (0*\horisep, -1*\vertsep+0pt) {$\phibf$};
        
        \draw[black!60] (x) -- (n1);
        \draw[black!60] (n1) -- (y);
        \draw[black!90] (z) -- (n2);
        \draw[black!90, line width=1.0pt] ([xshift=-0pt]n2.north) -- ([xshift=-0pt, yshift=6pt]n1.south);
        
        \draw [-, decorate,decoration={brace, amplitude=5pt,raise=0pt},yshift=0pt] 
            (-0.4*\horisep, 0.5*\vertsep) -- (0.4*\horisep, 0.5*\vertsep) 
            node [black,midway,yshift=5pt, above,align=center] {Primary model \\ $f_{\thetabf}(\x)$};
        \draw [-, decorate,decoration={brace, amplitude=5pt,raise=0pt},yshift=0pt] 
            (0.4*\horisep, -1.5*\vertsep) -- (-0.4*\horisep, -1.5*\vertsep) 
            node [black,midway,yshift=-5pt, below,align=center] {$g_{\phibf}(\z)$ \\ Implicit posterior model};
        
    \end{scriptsize}
\end{tikzpicture}
        \caption{Unconditioned posterior model.}
        \label{fig:modelArchitecture1}
    \end{subfigure}~
    \begin{subfigure}{0.45\columnwidth}
        \centering

\def\horisep{40pt}
\def\vertsep{25pt}
\def\horisepb{12pt}
\def\vertsepb{12pt}

\begin{tikzpicture}
    \begin{scriptsize}
        
        \tikzstyle{every path}=[->]
        
        \node[] (x) at (-1*\horisep, 0*\vertsep) {$\x$};
        \node[] (y) at (1*\horisep, 0*\vertsep) {$\y$};
        \node[] (x2) at (-1*\horisep, -0.85*\vertsep) {$\x$};
        \node[] (z) at (-1*\horisep, -1.15*\vertsep) {$\z$};
        \node[rectangle, fill=d61green!30, draw=black!60, minimum width=30pt, minimum height=18pt] (n1) at (0*\horisep, 0*\vertsep+0pt) {$\thetabf$};
        \node[rectangle, fill=d61blue!20, draw=black!60, minimum width=30pt, minimum height=18pt] (n2) at (0*\horisep, -1*\vertsep+0pt) {$\phibf$};
        
        \draw[black!60] (x) -- (n1);
        \draw[black!60] (n1) -- (y);
        \draw[black!90] (x2) -- (x2 -| n2.west);
        \draw[black!90] (z) -- (z -| n2.west);
        \draw[black!90, line width=1.0pt] ([xshift=-0pt]n2.north) -- ([xshift=-0pt, yshift=6pt]n1.south);
        
        \draw [-, decorate,decoration={brace, amplitude=5pt,raise=0pt},yshift=0pt] 
        (-0.4*\horisep, 0.5*\vertsep) -- (0.4*\horisep, 0.5*\vertsep) 
        node [black,midway,yshift=5pt, above,align=center] {Primary model \\ $f_{\thetabf}(\x)$};
        \draw [-, decorate,decoration={brace, amplitude=5pt,raise=0pt},yshift=0pt] 
        (0.4*\horisep, -1.5*\vertsep) -- (-0.4*\horisep, -1.5*\vertsep) 
        node [black,midway,yshift=-5pt, below,align=center] {$g_{\phibf}(\z,\x)$ \\ Implicit posterior model};
        
    \end{scriptsize}
\end{tikzpicture}
        \caption{Conditional posterior model.}
        \label{fig:modelArchitecture2}
    \end{subfigure}
    \caption{Architecture of the hierarchical Bayesian neural model, with input vectors $\x$ (predictor variables) and $\z$ (i.i.d. uniform random variates).  Subplot (a) shows the architecture where the implicit posterior is parametrically static and modelled as independent of $\x$ (the inputs to the primary model).  Subplot (b) shows the more flexible solution, where realisations from the implicit posterior model are conditioned on the input vector $\x$ with $\thetabf$ providing a local (rather than global) parameterisation of the primary model.  In both cases, the vector $\y$ is a vector that parameterises a likelihood function.  The parameters $\phibf$ of the overall model are optimised using gradient descent in order to maximise the posterior predictive distribution.}
    \label{fig:modelArchitecture}
\end{figure}

As illustrated in \figurename{~\ref{fig:modelArchitecture}}, we propose the use of two models: (1) a primary predictive model (e.g. a regressor or classifier), $f_{\thetabf}$, defined by parameter vector $\thetabf$, and (2) a posterior model $g_{\phibf}$ that generates random samples of the parameter vector $\thetabf$. A likelihood function is defined along with the primary model, and the posterior model approximates the (implicit) posterior distribution over the parameters of the primary model. The parameters of the primary model are indirectly trained by optimising (via gradient descent) the parameters, $\phibf$, of $g_{\phibf}$ in order to maximise the posterior predictive distribution.

The key contributions of this work are:
\begin{enumerate}
    \item We propose a simple and natural inference approach which optimises a posterior model according to the posterior predictive distribution. Other than the posterior model, our only approximation is a Monte Carlo estimate of the posterior predictive distribution which has an affinity to gradient descent optimisation, making it scalable to large datasets. Only a likelihood function (which is not limited to probability distributions) and the posterior model need be specified. The expressive implicit posterior model is the default, however parametric distributions are not excluded.
    \item We show that by modelling the conditional posterior (conditional on the primary model inputs), the apparent complexity of the primary model can be greatly increased. This increase is owing to the posterior model producing a unique set of parameters for any given input and therefore results in models with locally adaptive behaviour.
    \item We demonstrate that the implicit model used for the posterior distribution is highly flexible and capable of inducing multi-modality in the predictive distribution where appropriate.
\end{enumerate}

We see this work having impact in applications where uncertainty quantification is required such as medical, robotics, finance, and forecasting applications. The approach may also be useful in applications concerned with out-of-distribution data and problems relating to transfer learning. We see the conditional posterior model being highly applicable in problems relating to physics-based modelling, emulator/surrogate models, and anomaly detection.


\section{Methods}

\subsection{The Posterior Predictive Distribution Formulation}

The aim in a regression or classification problem is to train a model $f_{\thetabf}$ parametrised by $\thetabf$, using a dataset $\mathcal{D}$ = $\{(\x_1, \y_1), \dots, (\x_N, \y_N)\}$ of input-output pairs, such that, given a new input $\acute{\x} \notin \mathcal{D}$, the model is able to accurately predict the target $\y$ associated with $\acute{\x}$. In the Bayesian statistical approach, $\thetabf$ is treated as a \textit{latent} random variable and following the assimilation of observational data, our beliefs about which parameter values are acceptable can be updated (i.e. the posterior distribution). As such, a neural network modelled in this way is referred to as a Bayesian Neural Network (BNN) and can generate an ensemble of predictions $\acute{\y}_1, \dots, \acute{\y}_L$ for a given $\acute{\x}$ through the posterior predictive distribution. The overall challenge when using the Bayesian statistical approach is to find, estimate, or approximate the posterior distribution over $\thetabf$, which is given by Bayes' theorem $p(\thetabf | \mathcal{D}) = p(\mathcal{D}|\thetabf) p(\thetabf) / p(\mathcal{D})$. The marginal distribution in the denominator is generally intractable and approximate inference approaches are frequently required to infer the posterior.

In this study, we propose a simple approach whereby we attempt to directly infer the \textit{posterior predictive distribution}, which is the object of inference for making probabilistic predictions. Given a new input-output pair $(\acute{\x},\acute{\y}) \notin \mathcal{D}$, the posterior predictive distribution is given by \cite{gelman2013bayesian}
\begin{minipage}{\textwidth}
    \begin{align*}
        p(\acute{\y} | \acute{\x}, \mathcal{D}) 
        &= \int p(\acute{\y}, \thetabf |\acute{\x}, \mathcal{D}) d\thetabf \hspace{7cm}
    \end{align*}
    \vspace{-8pt}
    \begin{subnumcases}{\hspace{1.02cm} =}
        \int  p(\acute{\y} |\acute{\x}, \thetabf) p(\thetabf | \mathcal{D}) d\thetabf , & \textup{$\thetabf$ conditionally independent of $\acute{\x}$}, \label{eq:bayes_predictive_a} \\
        \int p(\acute{\y} | \acute{\x}, \thetabf) p(\thetabf| \acute{\x}, \mathcal{D}) d\thetabf , & \textup{$\thetabf$ conditionally dependent on $\acute{\x}$} \label{eq:bayes_predictive_b}
    \end{subnumcases}
    \vspace{3pt}
\end{minipage}
The annotations of (\ref{eq:bayes_predictive_a}) and (\ref{eq:bayes_predictive_b}) correspond to \figurename{} \ref{fig:modelArchitecture1} and \figurename{} \ref{fig:modelArchitecture2} and correspond to the situation where the posterior distribution is conditionally independent and dependent\footnote{(\ref{eq:bayes_predictive_b}) may appear like amortisation \cite{kingma2014autoencoding}, however the intention is to condition the posterior on $\acute{\x}$.} on $\acute{\x}$ respectively. We refer to these two models as (a) the \textit{unconditioned posterior model} and (b), the \textit{conditional posterior model}. See Appendix \ref{sec:posteriorPredictiveDistributionProof} for graphical models and a derivation.

The first factor in (\ref{eq:bayes_predictive_a}) and (\ref{eq:bayes_predictive_b}) is the likelihood of observing $\acute{\y}$ given $\acute{\x}$ and $\thetabf$. This likelihood can be evaluated via the elements that are output from the primary model $f_{\thetabf}(\acute{\x})$ and the targets $\acute{\y}$ from $\mathcal{D}$. The second factor is the posterior distribution given by Bayes theorem, which can be analytically intractable, or often only known up to a constant of proportionality. We propose to approximate the posterior with the distribution $q_{\phibf|\mathcal{D}}(\thetabf)$, which is parametrised by $\phibf$ and optimised with $\mathcal{D}$. The goal is then to find the parameters $\phibf$ that maximise the posterior predictive distribution:
\begin{align}
    \label{eq:optmisation_int}
    \phibf^* 
    & \triangleq \arg \max_{\phibf} \int p(\acute{\y} | \acute{\x}, \thetabf) q_{\phibf|\mathcal{D}}(\thetabf) d\thetabf.
\end{align}
Suppose we draw $L$ samples $\thetabf^{(1)}, \dots. \thetabf^{(L)}$ from $q_{\phibf|\mathcal{D}}(\thetabf)$. The Monte Carlo estimate of the posterior predictive distribution is
\begin{align}
    \label{eq:mc_bayes_predictive}
    p(\acute{\y} |\acute{\x}, \mathcal{D})
    &\simeq \frac{1}{L} \sum_{l=1}^L p(\acute{\y} |\acute{\x}, \thetabf^{(l)})
\end{align}
To optimise the model, the negative of (\ref{eq:mc_bayes_predictive}) can be used as a loss function. When we define the posterior model $g_{\phibf}(\z,\acute{\x}) \sim q_{\phibf|\mathcal{D}}(\thetabf)$ (where $\acute{\x}$ is included with (\ref{eq:bayes_predictive_b})), this loss function is given by
\begin{align}
    \label{eq:post_pred_loss}
    \mathcal{L} 
    &= - \frac{1}{L} \sum_{l=1}^L p(\acute{\y}| \acute{\x}, g_{\phibf}(\z^{(l)},\acute{\x}))
\end{align}
In practice, it may be useful to consider the negative log posterior estimate and make use of the log-sum-exp formulation for numerical stability.
The gradient of the loss function is
\begin{align}
    \frac{\partial \mathcal{L}}{\partial \phibf}
    &= - \frac{1}{L} \sum_{l=1}^L \frac{\partial p(\acute{\y}| \acute{\x}, g_{\phibf}(\z^{(l)},\acute{\x}))}{\partial \phibf} \frac{\partial g_{\phibf}(\z^{(l)},\acute{\x})}{\partial \phibf}
\end{align}
Note that, VI has the challenge of finding an unbiased, low variance Monte Carlo estimate of the ELBO's gradient \cite{mohamed2020monte}, whereas, in our approach, the target is a Monte Carlo estimate whose gradient is tractable. Given this gradient, the gradient-descent optimisation objective is then
\begin{align}
    \label{eq:optmisation}
    \phibf^* \triangleq \arg \min_{\phibf} \left[ - \frac{1}{L} \sum_{l=1}^L p(\acute{\y} | \acute{\x}, g_{\phibf}(\z^{(l)},\acute{\x})) \right]
\end{align}
A discussion on convergence in gradient descent optimisation for this form is provided in Appendix \ref{sec:convergenceProof}.

Following \figurename{~\ref{fig:modelArchitecture}} and equation (\ref{eq:optmisation}), $f_{\thetabf}$ is the primary model, intended to specify the parameters of a likelihood (or approximate likelihood) function $p(\acute{\y} | \acute{\x}, \thetabf)$.  For a sample of parameter vectors $\thetabf^{(1)}, \dots, \thetabf^{(L)}$, the loss function in (\ref{eq:post_pred_loss}) can be evaluated. The sampled vectors $\thetabf^{(l)}$ are generated by the approximate posterior model $g_{\phibf}(\z, \acute{\x})$ (parametrised by $\phibf$) that is intended to target $p(\thetabf | \mathcal{D})$.

\subsection{Primary Model and Likelihood}

The primary model $f_{\thetabf}$, that is responsible for parametrising the likelihood function, can conceivably be any model with parameter vector $\thetabf$.  Candidate primary models are parametrised functions that are differentiable with respect to their parameters. These include linear models, deep neural networks, and various physics-based models.

With an implicit posterior model, the likelihood associated with the primary model is not constrained to probability distributions. The likelihood operates as a loss or distance function, whose only requirements are those associated with gradient descent optimisation, such as convexity and differentiability \cite{boyd2004convex}. This is especially useful when a given loss function does not have a known associated density (or mass), such as heuristic loss functions, scaled errors in time-series forecasting, and the hinge losses in classification. Likelihood-free approaches such as approximate Bayesian Computation and synthetic likelihood approaches could also be adopted.

\subsection{Posterior Model}

The implicit posterior distribution is a generative process that transforms random samples $\z^{(l)}$ from some parametric distribution $\z$ with density $s(\cdot)$ (e.g. a uniform or standard normal) to samples from the posterior distribution $\thetabf^{(l)}$ according to a transform $g_{\phibf}(\z)$. If the transform $g_{\phibf}$ is a Borel-measurable function, $q_{\phibf|\mathcal{D}}(\thetabf)$ is a valid density given by \cite{mohamed2016learning,tran2017hierarchical,wasserman2013all}
\begin{align}
    q_{\phibf|\mathcal{D}}(\thetabf) = \frac{\partial}{\partial \theta_1} \cdots \frac{\partial}{\partial \theta_m} \int_{\{\z : g_{\phibf}(\z) \leq \thetabf \}} s(\z) d\z
\end{align}
The posterior model $g_{\phibf}(\z)$ can take various forms such as a neural network (which is sometimes referred to as a HyperNetwork \cite{ha2016hypernetworks}). As the posterior model models an implicit distribution, it makes no assumptions about the form of the posterior distribution, and only provides a means for sampling from it.

The posterior model described in \figurename{} \ref{fig:modelArchitecture1} operates by transforming samples from $\z$ to samples from the posterior. These generated samples can directly be applied in equation (\ref{eq:post_pred_loss}). In the conditional posterior model described in \figurename{} \ref{fig:modelArchitecture2}, $\acute{\x}$ is included as an input to the posterior model. Conceptually, this allows the posterior model to produce the best primary model parameters for the given input. The result is the apparent complexity of the primary model is increased. Conditioning on $\acute{\x}$ is similar to the idea of the conditional variational autoencoder (VAE) \cite{sohn2015learning} and the conditional generative adversarial network (GAN) \cite{mirza2014conditional}.

The posterior model does not exclude parametric distributions. A parametric distribution could be directly applied as an unconditioned posterior model, or a linear implicit model could be used. A mixture density network (MDN) \cite{bishop1994mixture} can be used to provide conditional posterior model. Similar to the architecture of the variational autoencoder's encoder \cite{kingma2014autoencoding}, the MDN transforms $\acute{\x}$ through a neural network to a posterior parametric mixture density, where samples can be drawn.

\subsection{Incorporating a Prior}
\label{sec:prior}

The prior $p(\thetabf)$ lurks within the posterior represented in (\ref{eq:bayes_predictive_a}) and (\ref{eq:bayes_predictive_b}). Without explicitly specifying it, we assume an improper, uninformative prior distribution that is constant across the domain of the parameters. This may be considered particularly convenient in context of black-box models such as neural networks when there is frequently no auxiliary information that can be drawn upon to formulate a prior.  Informative priors may also be difficult to represent given that such models often have high-dimensional parameter spaces and may exhibit complex patterns of dependence.

In some contexts, it may however be advantageous to include a prior. A weakly informative prior can be incorporated through regularisation. In fact, the objective of a prior is to regularize the posterior distribution, constraining it to contain the bulk of the probability mass within reasonable bounds \cite{gelman2013bayesian,jaynes2003probability}. Weakly informative priors may thus be enforced by (1) regularising the primary model with methods such as dropout (2) limiting the output of $g_{\phibf}$ to regularize $\thetabf$ to a specific range using an activation function, (3) specifying parametric distributions on the posterior and likelihood to force the prior into a family of distributions by conjugacy, (4) restricting the dimension of $\z$ to form a degenerate distribution on a union of manifolds within the domain of $\thetabf$ \cite{arjovsky2017towards}, and (5) restricting the complexity of $g_{\phibf}$ to regularize the posterior distribution complexity. 

Such regularisation approaches can however encourage degeneracy in the posterior as a degenerate distribution may be the simplest representation for the implicit model. In the extreme case, this may be a Dirac density (e.g. see Appendix \ref{sec:degeneratePosterior}).

\subsection{Model Training}
\label{sec:modelTraining}

To train the model, the parameters of $g_{\phibf}$ are optimised via gradient descent such that $g_{\phibf}$ produces a distribution over $\thetabf$ that maximises the posterior predictive distribution. The parameters of the primary model $\thetabf$ are thus indirectly optimised by optimising $\phibf$. The algorithm for training the model is provided in Algorithm \ref{alg:training}. The use of minibatches is discussed in Appendix \ref{sec:minibatches}.

\begin{algorithm}[!b]
    \begin{scriptsize}
        \begin{algorithmic}[]
            \Require primary model $f_{\thetabf}$, posterior model $g_{\phibf}$, training data $\mathcal{D}$, minibatch size $B$, number of MC samples $L$, number of epochs $K$.
            \State Initialise the posterior model parameters $\phibf$ randomly.
            \For{$i \in (1,K)$} 
            \For{minibatch $n \in \mathcal{D}$}
            \For{$l \in (1,L)$} 
            \State Sample $\z$ (depending on the architecture of $g_{\phibf}$).
            \State Sample $\thetabf^{(l)}$ from $\g_{\phibf}(\z)$. 
            \State Pass $\x_n$ through $f(\thetabf^{(l)})$ to produce $\hat{\y}_n$.
            \State Evaluate the likelihood $p(\y_n |\x_n, \thetabf^{(l)})$ given $\hat{\y}_n$.
            \EndFor
            \State Compute the loss $\mathcal{L}_n = - \frac{1}{L} \sum_{l=1}^L p(\y_n | \x_n, \thetabf^{(l)})$.
            \State Compute the gradients $\frac{\partial \mathcal{L}_n}{\partial \phibf}$.
            \State Backpropagate and update $\phibf$. 
            \EndFor
            \EndFor
            \State \Return $\phibf$
        \end{algorithmic}
    \end{scriptsize}
    \caption{Model training.}
    \label{alg:training}
\end{algorithm}

Given that the Monte Carlo estimate of the posterior predictive distribution in (\ref{eq:mc_bayes_predictive}) is based on the likelihood function and that an uninformed prior is assumed, it is easy to see that the approach has the potential to converge towards a degenerate distribution at the maximum likelihood (see Appendix \ref{sec:degeneratePosterior}). To avoid this, the use of early stopping is essential and is used in all experiments. 

Note that the posterior model has been described as a model over the complete set of $\thetabf$. However, in practice it may be convenient to define a separate posterior model for subsets of $\thetabf$. For example, if $f_{\thetabf}$ is a neural network, a separate posterior model can be defined for each layer in $f_{\thetabf}$. This reduces memory capacity as it assumes independence of the parameters across layers and it is more convenient to implement in deep learning frameworks.


\section{Results}

We demonstrate initial results relating to our contributions described in Section \ref{sec:introduction}. The aim is to demonstrate our approach and its features. For this, a pedagogical comparison is taken to highlight differences between various approaches and provoke thought. A detailed performance comparison (e.g. uncertainty coverage) across multiple datasets and models is reserved for future work.

To provide effective demonstrations, synthetic datasets are constructed with specific properties. We also test our approach in a forecasting problem with a larger dataset and more complex primary model in Section \ref{sec:forecastingDemonstration}.

Additional results are provided in Appendices \ref{sec:posteriorDistributionModelConfigurations} and \ref{sec:priorInclusionDemonstrationAppendix}.

\subsection{Bayesian Inference with Uncertainty Estimates}
\label{sec:bayesianInferenceWithUncertaintyEstimates}

We consider a synthetic dataset of containing 32 samples from the function $y = x \sin(x)$ and additional high-variance samples \{(7,-7), (8.5,7), (10,-7), (11.5,7)\} \cite{gal2016uncertainty}.  The additional high-variance samples (originally intended to emulate a heteroscedastic process) are used to demonstrate how the different models adapt to structural variations in the data. The test dataset contains input samples at a finer resolution than the training dataset (1024 samples). Furthermore, the range of the test inputs extends beyond the range of the training set to provide of out-of-distribution samples. 

\begin{figure}[!tb]
    \centering
    \includegraphics{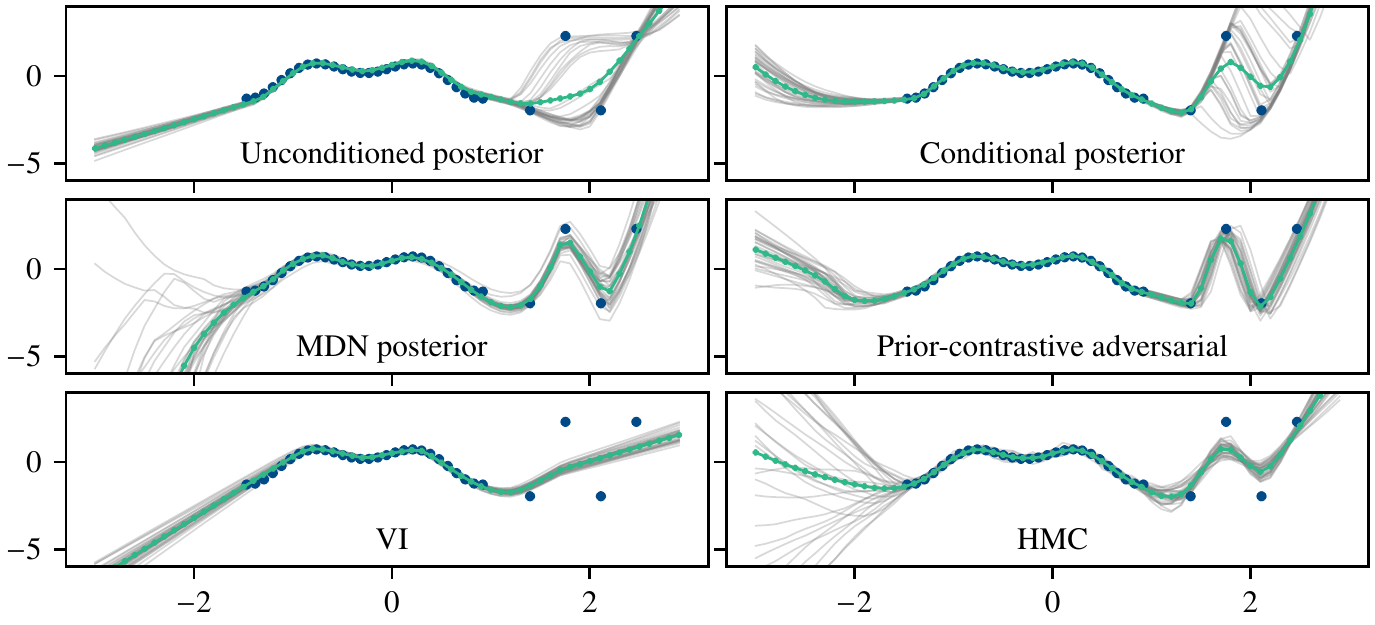}
    \caption{Results for the MLP primary model. Blue markers are the targets, the grey curves are 30 MC samples from the posterior predictive distribution, and the green curve is the mean of these samples. Note that these figures plot standardised data.}
    \label{fig:results}
\end{figure}

We demonstrate the proposed approach and compare it with a VI approach (Bayes by Backprop) \cite{blundell2015Weight}, the HMC approach \cite{neal2011mcmc}, and the prior-contrastive adversarial VI approach \cite{huszar2017variational,mescheder2017adversarial}.  For the proposed approach, we consider conditional, unconditioned, and MDN posterior models. In all cases, a Multilayer-Perceptron (MLP) serves as a regression model and the likelihood is assumed to be a Gaussian with a known standard deviation of 0.01. The models are trained with 10 Monte Carlo samples and 30 Monte Carlo samples are drawn from the posterior predictive distribution during testing. Model configurations are described in Table \ref{table:modelDetails} and the results are illustrated in \figurename{~\ref{fig:results}}. 

\begin{table}[t]{}
    \caption{Model architecture details.}
    \label{table:modelDetails}
    \begin{center}
        \begin{scriptsize}
            \setlength{\tabcolsep}{5pt}
            \begin{tabular}{l p{0.80\columnwidth}}
                \toprule
                Model & Architecture \\
                \midrule
                Primary model & MLP with a hidden layer comprising 512 ReLU hidden units and a linear output. \\
                Conditional Posterior &  MLP with 4 uniform random inputs with the input $\x$ and 16 ReLU hidden units. \\
                Unconditioned Posterior &  MLP with 4 uniform random inputs and 16 ReLU hidden units. \\
                MDN Posterior &  MLP with the input $\x$, 16 ReLU hidden units, and unimodal Gaussians at the output. \\
                VI &  Mean field variational assumption with a spike and slab priors as in \cite{blundell2015Weight} \\
                HMC & Factorized Gaussian prior. The algorithm is run over 4000 steps with 25 leapfrog steps with step size 0.001. \\
                Prior-contrastive & Generative model equivalent to the conditional posterior model and a MLP discriminator with 512 ReLU hidden units. \\
                \bottomrule
            \end{tabular}
        \end{scriptsize}
    \end{center}
\end{table}
%

%

All models treat the high-variance samples differently. The conditional posterior model seems to embrace the change in the structure of the data as its variance expands to include these samples. The unconditioned model appears to have discovered two possible paths through the high-variance data samples, suggesting a bimodal posterior. The MDN and prior-contrastive adversarial models tend to provide a close fit the high-variance samples, whereas VI seems to have treated these samples as outliers. HMC seems to partially fit to these samples, but its variance remains relatively low.

The variance of the models all expand over the out-of-distribution regions to some degree. HMC has a rapidly expanding variance in the out-of-distribution region to the left of the data. The MDN model also has a rapidly expanding variance, however its mean is significantly different to that of HMC. The variance of the conditional posterior model and the prior-contrastive adversarial model increase in the out-of-distribution regions, but not as significantly as HMC. VI generally seems to underestimate the variance overall as expected \cite{bishop2006pattern}.

\subsection{Increased Model Complexity}

\begin{figure}[!tb]
    \centering
    \includegraphics{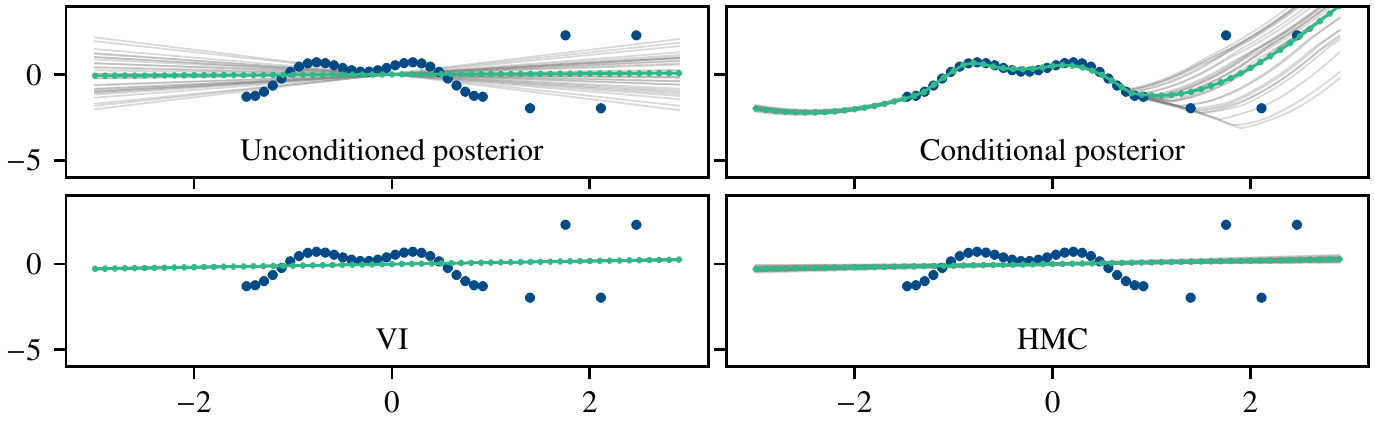}
    \caption{Results with a linear regression model for the primary model. Blue markers are the targets, the grey curves are 30 MC samples from the posterior predictive distribution, and the green curve is the mean of these samples. Note that these figures plot standardised data.}
    \label{fig:resultsLr}
\end{figure}

We consider testing the posterior models with a simple linear regression primary model (a straight line) on the non-linear $y = x \sin(x)$ dataset described in Section \ref{sec:bayesianInferenceWithUncertaintyEstimates}. This allows us to assess the affect of the conditional posterior model on the primary model. The results are presented in \figurename{~\ref{fig:resultsLr}}. 

%
%
\begin{figure}[!tb]
    \centering
    \includegraphics{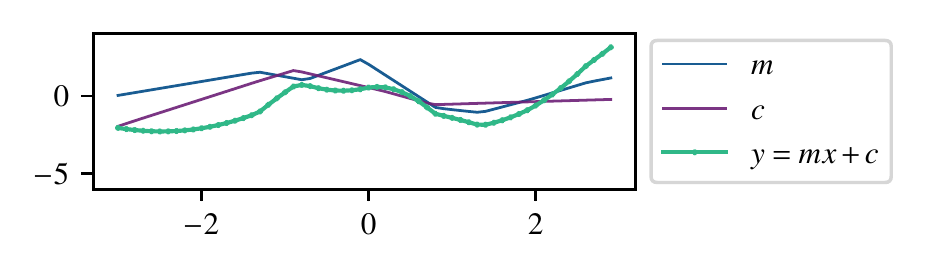}
    \caption{A plot of a sample of primary model parameters and the prediction with these parameters. With a linear primary model, the parameters correspond to a gradient and offset of a straight line. The key observation is that the parameters vary over $\x$ to produce a non-linear prediction.}
    \label{fig:results_weights}
\end{figure}

What is of interest is that the conditional posterior approach fits the non-linear data with a \textit{linear} model. In comparison, the other models fit the data with straight lines as expected. The conditional posterior model has learned to localise the parameters of the primary model and vary them according to the input $\acute{\x}$. The parameters $\thetabf$ are varied over the continuous space of $\acute{\x}$ as illustrated in \figurename{~\ref{fig:results_weights}}. This result is similar to local regression methods \cite{loader2006local,hastie2013elements}, but over a continuous range. A reasonable consequence is that a simpler primary model can be used, reducing overall complexity of the approach.

\subsection{Likelihood Flexibility}
\label{sec:likelihoodFlexibilityResults}

The posterior distribution is only constrained by the likelihood function and could, for example, be multimodal. A multimodal posterior would suggest several variations of parameters corresponding to several output options for a given input. For example, this can occur in tracking applications where targets can split and merge, or inverse problems with one-to-many mappings. 

To test this, a synthetic dataset based on $x \sin(x)$ (with scaled inputs) is generated according to
\begin{align}
    y =
    \begin{cases}
        a(x) \sin(a(x)) + \epsilon & x \in (0,0.6) \\
        a(x) \sin(a(x)) + 1 + \epsilon & x \in (0.3,1)
    \end{cases}
\end{align}
where $\epsilon \sim \mathcal{N}(0, 0.01)$ and $a(x) = 10x-5$ scales the inputs to the range (0,1). The key feature of this function is that there are two parallel curves over the domain $(0.3, 0.6)$. A total of 128 samples are drawn from this function ensuring that each curve has the same number of samples. Ideally, the model should predict the lower curve over $(0,0.3)$, either curves over $(0.3, 0.6)$, and the upper curve over $(0.6, 1)$.

\begin{figure}[!tb]
    \centering
    \includegraphics{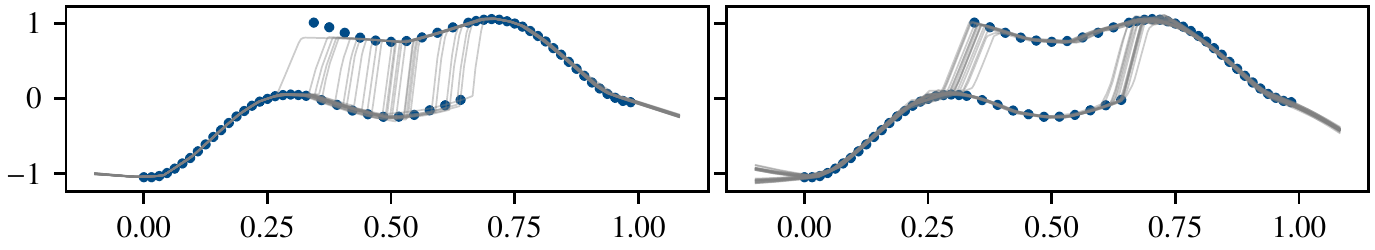}
    \caption{Model predictions for a multimodal dataset. Left: using the $L1$-norm as a likelihood function. Right: including sample labels as inputs to the posterior model. Blue markers are the targets and the grey curves plot 30 Monte Carlo prediction samples. The posterior model is bimodal over the domain $(0.3, 0.6)$ where it can predict either curve, and is unimodal otherwise.}
    \label{fig:multimodal}
\end{figure}

To achieve varying parameters over $\x$, the conditional posterior model is used. To encourage a bimodal posterior distribution, we consider using the $L1$-norm as a likelihood function. Note that the $L1$-norm is not a density function, highlighting the flexibility of the likelihood specification. To support non-smooth transitions between curves, the primary model should use non-smooth activation functions, such as the ReLU. The results are illustrated in left panel of \figurename{~\ref{fig:multimodal}}. The MC samples from the model (grey curves), illustrate how the model is able to switch between the different curves over the domain $(0.3, 0.6)$. According to the $L1$-norm, both parallel curves have the same likelihood and the model is permitted to switch between them at any time. Owing to the conditioning on $\acute{\x}$, the model reverts to a unimodal posterior and predicts one of the appropriate curves outside of the domain $(0.3, 0.6)$.
%

If there are labels indicating which curve the targets originate from, these labels could be included as inputs to the posterior model along with $\acute{\x}$. These labels provide an indication to the posterior model of a potential multimodal posterior distribution. To demonstrate this, a label is provided as an input to the conditional posterior model and a unimodal Gaussian likelihood is used. The labels are generated by labelling even indexed samples with zeros and odd indexed samples with ones in the domain $(0.3, 0.6)$. During testing, the labels are randomly assigned. The results are illustrated in the right panel of \figurename{} \ref{fig:multimodal}. The model learns a clearer distinction between the curves compared with the $L1$-norm-based approach. Note that a bimodal posterior is learned despite a unimodal likelihood.

\subsection{Forecasting Demonstration}
\label{sec:forecastingDemonstration}

The proposed approach is applied to the N-BEATS model \cite{oreshkin2020nbeats}
in a multi-step-ahead forecasting task. N-BEATS is used as the primary model, and we compare results with the unconditioned posterior model, the conditional posterior model, and standard maximum likelihood training. The dataset used is the monthly average temperature in England over the period of January 1723 to December 1970 from Rob Hyndman's Time-Series Data Library\footnote{Rob Hyndman's Time-Series Data Library: \url{https://pkg.yangzhuoranyang.com/tsdl/}. LIC: GPL-3.}. The model is configured to forecast 3 months ahead given the last 6 months of data. Input-output pairs are created with a 6-sample input window and its adjacent 3-sample output window, which are slid across the dataset. A total of 2960 input-output pairs are formed, which are split into a training and testing set comprising 2075 and 885 samples respectively. The model architecture details are provided in Table \ref{table:forecastModelDetails}. The models are trained over 100 epochs using ADAM with a learning rate of 0.01 and a batch size of 128.

\begin{table}[t]{}
    \caption{Model architecture details for the forecasting results.}
    \label{table:forecastModelDetails}
    \begin{center}
        \begin{scriptsize}
            \setlength{\tabcolsep}{5pt}
            \begin{tabular}{l p{0.80\columnwidth}}
                \toprule
                Model & Architecture \\
                \midrule
                N-BEATS &  Input sequence length: 24, output sequence length: 12, stacks: 1, blocks: 3, block type: generic $\theta$-dimension: 32, FC stack layer dimension: 64, block weights: shared. \\
                Posterior models &  MLP with 4 uniform random inputs (and the input $\x$ for the conditional model model) and 16 hidden neurons with ReLU activation functions. \\
                \bottomrule
            \end{tabular}
        \end{scriptsize}
    \end{center}
\end{table}

The forecasting results are provided in Table \ref{table:forecastResults} in terms of Root Mean Squared Error (RMSE), Mean Absolute Percentage Error (MAPE), and the epoch run time. As a baseline results for Na\"{\i}ve forecasting method are also provided. The results demonstrate that the standard N-BEATS model produces the highest errors and the N-BEATS model with a conditional posterior model produces the lowest errors. N-BEATS is able to produce lower errors owing to the increased modelling capacity through the conditional posterior distribution. With the unconditioned posterior model, N-BEATS is able to generalise better than the standard training owing to less overfitting, where Bayesian models are more inclined to find ``flat minima'' that provide better generalisation \cite{wilson2020bayesian}.

The epoch times demonstrate an increased computational complexity of the proposed approach compared to the standard maximum likelihood training. This increase is due to the additional complexity of the posterior model architecture and the multiple MC samples. This cost however comes with the benefits of model uncertainty and improved generalisation. Furthermore, memory requirements could be reduced to linear scaling by using weight normalization reparameterisation and only outputting the scaling factors from the posterior model \cite{krueger2017bayesian}.

\begin{table}[t]{}
    \caption{Average errors over the test dataset. Epoch times are provided for a 2.6 GHz 6-Core Intel Core i7 with 16GB RAM. Note that the relatively high epoch time for the conditional posterior model is due to the minibatch strategy (see Appendix \ref{sec:minibatches}).}
    \label{table:forecastResults}
    \begin{center}
        \begin{scriptsize}
            \setlength{\tabcolsep}{5pt}
            \begin{tabular}{l c c c c c}
                \toprule
                Model & RMSE & RMSE std dev. & MAPE & MAPE std dev. & Epoch run time (s) \\
                \midrule
                Naive									& 4.47 & 2.13 & 72.5 & 135.0 & N/A \\
                N-BEATS 								& 2.87 & 1.11 & 55.6 & 156.9 & 0.22 \\
                N-BEATS with unconditioned posterior 	& 1.86 & 0.76 & 35.7 & 101.9 & 1.31 \\
                N-BEATS with conditional posterior 		& 1.35 & 0.64 & 27.8 & 94.9  & 4.2 \\
                \bottomrule
            \end{tabular}
        \end{scriptsize}
    \end{center}
\end{table}
%


\section{Conclusion}

We have proposed a novel approach for Bayesian inference in complex models and demonstrated key features of the approach. We show how the approach is able to provide uncertainty estimates in high variance samples and out-of-distribution regions, model multi-modal posterior distributions, improve capacity of a primary model, and provide better generalisation in a forecasting task.

The approach does not rely on MCMC or VI and provides flexibility in the likelihood specification. The model targets the posterior predictive distribution in training, which is natural as this is generally the primary distribution of interest in practice. Training is performed via gradient descent, providing scalability to large datasets.

Future work involves applying the approach to the domains described in Section \ref{sec:introduction}, especially the physics-based and surrogate modelling domains where we see the conditional posterior approach having impact.

\section*{Acknowledgements}

We would like to thank Edwin Bonilla for the various discussions around this topic of this paper. This work was supported by the CSIRO MLAI Future Science Platform.

    \bibliographystyle{unsrtnat}
    \bibliography{Bibliography}
    
    \clearpage

    \appendix
\section{Posterior Predictive Distribution Proof}
\label{sec:posteriorPredictiveDistributionProof}

The proof of (\ref{eq:bayes_predictive_a}) and (\ref{eq:bayes_predictive_b}) is provided using $d$-separation:

\begin{proof}
    Consider the dataset input-output pairs given by $\mathcal{D}$ = $\{(\x_i, \y_i)\}_{i=1}^N$ and a primary model $f_{\thetabf}(\x)$ parameterised by $\thetabf$. 
    The conditional independence assumptions are illustrated in the graphical model presented in \figurename{~\ref{fig:fig_bayesianGraphicalModel}}. 
    The posterior predictive distribution for a new input-output pair $(\acute{\x},\acute{\y}) \notin \mathcal{D}$ is given by
    \begin{align*}
        p(\acute{\y} | \acute{\x}, \mathcal{D})
        &= \int p(\acute{\y}, \thetabf |\acute{\x}, \mathcal{D}) d\thetabf
    \end{align*}
    By the definition of conditional probability
    \begin{align}
        p(\acute{\y} | \acute{\x}, \mathcal{D})
        &= \int p(\acute{\y} | \thetabf, \acute{\x}, \mathcal{D}) p(\thetabf| \acute{\x}, \mathcal{D}) d\thetabf
    \end{align}
    In the first factor, $\acute{\y}$ are independent of $\mathcal{D}$ given that the inputs are i.i.d and given that the outputs are conditionally independent given $\thetabf$, which produces
    \begin{align}
        \label{eq:posteriorPredictiveProof1}
        p(\acute{\y} | \acute{\x}, \mathcal{D})
        &= \int p(\acute{\y} | \acute{\x}, \thetabf) p(\thetabf| \acute{\x}, \mathcal{D}) d\thetabf
    \end{align}
    This result corresponds to the conditional posterior model described in equation (\ref{eq:bayes_predictive_b}) and \figurename{} \ref{fig:modelArchitecture2}. In the the second factor, $\thetabf$ is conditioned on $\acute{\x}$ as indicated by the dashed link in \figurename{~\ref{fig:fig_bayesianGraphicalModel}}. If this conditioning were removed, $\thetabf$ is conditionally independent on $\acute{\x}$ given that $\acute{\y}$ is not observed. Thus
    \begin{align}
        \label{eq:posteriorPredictiveProof2}
        p(\acute{\y} | \acute{\x}, \mathcal{D})
        &= \int p(\acute{\y} | \acute{\x}, \thetabf) p(\thetabf | \mathcal{D}) d\thetabf
    \end{align}
    This result corresponds to the unconditional posterior model described in equation (\ref{eq:bayes_predictive_a}) and \figurename{} \ref{fig:modelArchitecture1} (and is the common representation of the posterior predictive distribution in the literature).
\end{proof}
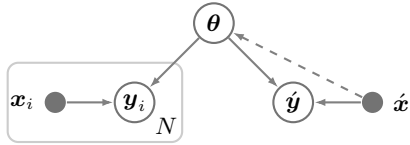
\begin{figure}[!h]
    \centering

\def\horisep{30pt}
\def\vertsep{30pt}

\begin{tikzpicture}[->,draw=black!50, thick]
    \begin{small}
        \tikzstyle{var}=[circle,fill=black!55,minimum size=8pt,inner sep=0pt]
        \tikzstyle{rv}=[circle,draw=black!55,minimum size=15pt,inner sep=1pt]

        \draw[rounded corners, draw=black!20] (-0.6*\horisep, -0.5*\vertsep) rectangle (1.6*\horisep, 0.5*\vertsep) {};
        \node[] at (1.4*\horisep, -0.3*\vertsep){$N$};
        
        \node[var, label=left:$\x_i$] (x) at (0*\horisep, 0*\vertsep){};
        \node[rv] (y) at (1*\horisep, 0*\vertsep) {$\y_i$};
        \node[rv] (w) at (2*\horisep, 1*\vertsep) {$\thetabf$};
        
        \node[var, label=right:$\acute{\x}$] (xp) at (4*\horisep, 0*\vertsep){};
        \node[rv] (yp) at (3*\horisep, 0*\vertsep) {$\acute{\y}$};

        \path[] (x) edge (y);
        \path[] (w) edge (y);
        \path[] (xp) edge (yp);
        \path[] (w) edge (yp);
        \path[dashed] (xp) edge (w);
        
    \end{small}
\end{tikzpicture}
    \caption{Graphical model for general Bayesian modelling. The data $\mathcal{D}$ = $\{(\x_1, \y_1), \dots, (\x_N, \y_N)\}$  are assumed i.i.d. and represented using plate notation, the inputs $\x$ are not treated as random variables, and the outputs $\y$ are assumed conditionally independent given the model parameters $\thetabf$. The dashed line corresponds to the conditional posterior model in \figurename{~\ref{fig:modelArchitecture2}}, where $\thetabf$ is conditioned on $\acute{\x}$.}
    \label{fig:fig_bayesianGraphicalModel}
\end{figure}
%


\section{Convergence}
\label{sec:convergenceProof}

Our approach is to encapsulate the Bayesian inference problem within a frequentist optimisation problem. In this frequentist approach, we treat the inputs $\z$ and $\x$, and the targets $\y$ as random, and we use them to find a point estimate of $\phibf$. As described in (\ref{eq:optmisation_int}), this optimisation results in inferring the posterior distribution of $\thetabf$ given $\mathcal{D}$. 

In this context, our approach conforms to the standard stochastic gradient descent (SGD) problem and thus converges accordingly. That is, random samples (or batches) of inputs ($\z$ and $\x$) and the targets ($\y$) are used to maximise $\phibf$ according to a ``likelihood'' given by (\ref{eq:post_pred_loss}). Various proofs and discussions on the convergence of SGD are available in the literature, such as in  \citet[Ch. 6]{bubeck2015convex}. 

Note that, although (\ref{eq:post_pred_loss}) is treated as a likelihood in the encapsulating frequentist problem, it is the MC estimate of the posterior predictive distribution which directs the Bayesian inference problem. By optimising $\phibf$ with maximum likelihood, we indirectly infer the posterior distribution of $\thetabf$ given $\mathcal{D}$.

The maximum likelihood optimisation approach provides insight into the requirements for the likelihood function $p(\acute{\y} |\acute{\x}, \thetabf^{(l)})$. In general, convergence of SGD requires that the objective $\mathcal{L}$ is convex and differentiable. Suppose $-p(\acute{\y}| \acute{\x}, \thetabf^{(l)})$ is given by some distance or loss function $c(\acute{\y}, \hat{\y}^{(l)})$ (e.g. mean squared error), where $\hat{\y}^{(l)} = f(\acute{\x}; \thetabf^{(l)})$ and $\thetabf^{(l)} = g_{\phibf}(\z^{(l)})$. The loss function in (\ref{eq:post_pred_loss}) is then given by
\begin{align}
    \mathcal{L} 
    &= - \frac{1}{L} \sum_{l=1}^L p(\acute{\y}| \acute{\x}, g_{\phibf}(\z^{(l)})) \nonumber \\
    &= \frac{1}{L} \sum_{l=1}^L c(\acute{\y}, f(\acute{\x}; g_{\phibf}(\z^{(l)})))
\end{align}
Given that the sum of convex functions is also convex, if $c$ is convex then $\mathcal{L}$ is convex (e.g. if $c$ is convex, $\nabla_{\phibf}^2 c_l \geq 0 \implies \sum_{l=1}^L \nabla_{\phibf}^2 c_l \geq 0 \implies \sum_{l=1}^L c_l$ is convex). As a result, $c$ can generally be any distance or cost function that is used in machine learning literature, where a key requirement is differentiability. Note however that, as with training standard neural networks, if $f$ and $g$ are non-linear. $c$ may be non-convex and the solution is only locally optimal. 

\section{Degenerate Implicit Posterior}
\label{sec:degeneratePosterior}

A proof by construction (or proof by example) approach is taken to show that the implicit posterior can produce a degenerate distribution which is located at the maximum likelihood, as described in sections \ref{sec:prior} and \ref{sec:modelTraining}. (Note that the selected example makes use of the unconditioned posterior model for notational convenience. The example is however trivially extended to the conditional posterior model case, which includes both $\z$ and $\x$ as inputs.)

\begin{proof}
    Consider an unconditioned posterior model in the form of an MLP with $K$ layers, weights $\W_{1:M}$ biases $\b_{1:M}$, and activation functions $\psi()$:
    \begin{align}
        \thetabf = (\psi_{\W_K, \b_K} \circ \psi_{\W_{K-1}, \b_{K-1}} \circ \cdots \circ \psi_{\W_{2}, \b_{2}} \circ \psi_{\W_1, \b_1})(\z)
    \end{align}
    If $\W_1 = \mathbf{0}$
    \begin{align}
        \psi_{\W_1, \b_1}(\z) = \psi(\W_1 \z + \b_1) = \psi(\mathbf{0} \z + b_1) = \psi(\b_1)
    \end{align}
    The inputs $\z$ are ignored by the MLP resulting in a constant output with respect to $\z$ (the same result is achieved in the conditional posterior model when the weights associated with both $\z$ and $\x$ are zero). The parameters $\phibf$ however can still be optimised to provide an optimal, but fixed output for $\thetabf$, denoted by $\tilde{\thetabf}$. 
    
    Consider the objective function in (\ref{eq:post_pred_loss})
    \begin{align}
        \mathcal{L} 
        &= - \frac{1}{L} \sum_{l=1}^L p(\acute{\y}| \acute{\x}, g_{\phibf}(\z^{(l)}))
    \end{align}
    If $g_{\phibf}$ is deterministic and fixed 
    \begin{align}
        \tilde{\thetabf} 
        = g_{\phibf}(\z^{(i)}) 
        = g_{\phibf}(\z^{(j)}) ~ \forall ~ i,j \in (1, \dots, L)
    \end{align}
    and thus
    \begin{align}
        p(\acute{\y}| \acute{\x}, \tilde{\thetabf}) 
        = p(\acute{\y}| \acute{\x}, g_{\phibf}(\z^{(i)})) 
        = p(\acute{\y}| \acute{\x}, g_{\phibf}(\z^{(j)}))  ~ \forall ~ i,j \in (1, \dots, L)
    \end{align}
    The MC estimate of the posterior predictive distribution is then
    \begin{align}
        \mathcal{L} 
        = -\frac{1}{L} \sum_{l=1}^L p(\acute{\y}| \acute{\x}, \tilde{\thetabf})) 
        = -p(\acute{\y}| \acute{\x}, \tilde{\thetabf})
    \end{align}
    which is the negative likelihood of $\acute{\y}$ given $\tilde{\thetabf}$. Minimising $g_{\phibf}$ produces the optimal parameters for $f_{\thetabf}$ according to this likelihood, which produces a maximum likelihood estimate.
\end{proof}

From the Bayesian perspective, the proved result is a degenerate Dirac distribution that is located at the maximum likelihood solution. Note that with an uninformed prior, the maximum likelihood estimate is equivalent to the maximum a posteriori estimate.
\section{Minibatches}
\label{sec:minibatches}

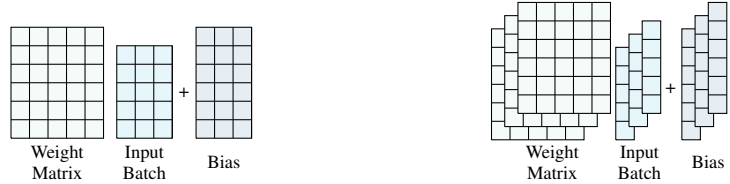
\begin{figure}[!t]
    \centering

\def\elmtsize{7pt}
\def\offset{5pt}

\begin{tikzpicture}
    \begin{scriptsize}
        \tikzstyle{block}=[rectangle, draw=black!40, fill=d61green!5, minimum width=1.8cm, minimum height=9pt, inner sep=3pt, align=center]
        \tikzstyle{circ}=[circle, draw=black!50, fill=d61green!50, minimum size=10pt, inner sep=3pt, align=center]
        \tikzstyle{dots}=[circle, inner sep=0pt, align=center]
        \tikzstyle{outp}=[circle, draw=black!50, fill=d61green!20, minimum size=15pt, inner sep=0pt, align=center]

        \draw[fill=d61green!5] (0,0) -- (5*\elmtsize,0) -- (5*\elmtsize,6*\elmtsize) -- (0, 6*\elmtsize) -- (0,0);
        \foreach \x in {1,...,5}
        {
            \draw[] (0,\x * \elmtsize) -- (5*\elmtsize, \x * \elmtsize);
        }
        \foreach \x in {1,...,4}
        {
            \draw[] (\x * \elmtsize,0) -- (\x * \elmtsize, 6*\elmtsize);
        }

        \draw[fill=d61blue!10] (5*\elmtsize+\offset,0) -- (8*\elmtsize+\offset,0) -- (8*\elmtsize+\offset,5*\elmtsize) -- (5*\elmtsize+\offset, 5*\elmtsize) -- (5*\elmtsize+\offset,0);
        \foreach \x in {1,...,4}
        {
            \draw[] (5*\elmtsize+\offset,\x * \elmtsize) -- (8*\elmtsize+\offset, \x * \elmtsize);
        }
        \foreach \x in {1,...,3}
        {
            \draw[] (\x * \elmtsize+5*\elmtsize+\offset,0) -- (\x * \elmtsize+5*\elmtsize+\offset, 5*\elmtsize);
        }
    
        \node[]() at (9*\elmtsize + 0.5*\offset, 2.5*\elmtsize) {+};
        
        \draw[fill=d61oceanblue!10] (10*\elmtsize,0) -- (13*\elmtsize,0) -- (13*\elmtsize,6*\elmtsize) -- (10*\elmtsize, 6*\elmtsize) -- (10*\elmtsize,0);
        \foreach \x in {1,...,5}
        {
            \draw[] (10*\elmtsize,\x * \elmtsize) -- (13*\elmtsize, \x * \elmtsize);
        }
        \foreach \x in {1,...,3}
        {
            \draw[] (\x * \elmtsize+10*\elmtsize,0) -- (\x * \elmtsize+10*\elmtsize, 6*\elmtsize);
        }
    
        \node[align=center] () at(2.5*\elmtsize, -1.3*\elmtsize) {Weight\\Matrix};
        \node[align=center] () at(6.5*\elmtsize+\offset, -1.3*\elmtsize) {Input\\Batch};
        \node[align=center] () at(11.5*\elmtsize, -1.3*\elmtsize) {Bias};

    \end{scriptsize}
\end{tikzpicture}
\hspace{3cm}
\begin{tikzpicture}
    \begin{scriptsize}
        \tikzstyle{block}=[rectangle, draw=black!40, fill=d61green!5, minimum width=1.8cm, minimum height=9pt, inner sep=3pt, align=center]
        \tikzstyle{circ}=[circle, draw=black!50, fill=d61green!50, minimum size=10pt, inner sep=3pt, align=center]
        \tikzstyle{dots}=[circle, inner sep=0pt, align=center]
        \tikzstyle{outp}=[circle, draw=black!50, fill=d61green!20, minimum size=15pt, inner sep=0pt, align=center]

        \foreach \y in {1,...,3}
        {
            \draw[fill=d61green!5] (0+\y*\offset,0+\y*\offset) -- (5*\elmtsize+\y*\offset,0+\y*\offset) -- (5*\elmtsize+\y*\offset,6*\elmtsize+\y*\offset) -- (0+\y*\offset, 6*\elmtsize+\y*\offset) -- (0+\y*\offset,0+\y*\offset);
            \foreach \x in {1,...,5}
            {
                \draw[] (0+\y*\offset,\x * \elmtsize+\y*\offset) -- (5*\elmtsize+\y*\offset, \x * \elmtsize+\y*\offset);
            }
            \foreach \x in {1,...,4}
            {
                \draw[] (\x * \elmtsize+\y*\offset,0+\y*\offset) -- (\x * \elmtsize+\y*\offset, 6*\elmtsize+\y*\offset);
            }
        
            \draw[fill=d61blue!10] (6*\elmtsize+\offset+\y*\offset,0+\y*\offset) -- (7*\elmtsize+\offset+\y*\offset,0+\y*\offset) -- (7*\elmtsize+\offset+\y*\offset,5*\elmtsize+\y*\offset) -- (6*\elmtsize+\offset+\y*\offset, 5*\elmtsize+\y*\offset) -- (6*\elmtsize+\offset+\y*\offset,0+\y*\offset);
            \foreach \x in {1,...,4}
            {
                \draw[] (6*\elmtsize+\offset+\y*\offset,\x * \elmtsize+\y*\offset) -- (7*\elmtsize+\offset+\y*\offset, \x * \elmtsize+\y*\offset);
            }

            \draw[fill=d61oceanblue!10] (11*\elmtsize+\y*\offset-\offset,0+\y*\offset) -- (12*\elmtsize+\y*\offset-\offset,0+\y*\offset) -- (12*\elmtsize+\y*\offset-\offset,6*\elmtsize+\y*\offset) -- (11*\elmtsize+\y*\offset-\offset, 6*\elmtsize+\y*\offset) -- (11*\elmtsize+\y*\offset-\offset,0+\y*\offset);
            \foreach \x in {1,...,5}
            {
                \draw[] (11*\elmtsize+\y*\offset-\offset,\x * \elmtsize+\y*\offset) -- (12*\elmtsize+\y*\offset-\offset, \x * \elmtsize+\y*\offset);
            }

        }
        \node[]() at (10.4*\elmtsize, 3.5*\elmtsize) {+};
        
        \node[align=center] () at(4*\elmtsize, -0.5*\elmtsize) {Weight\\Matrix};
        \node[align=center] () at(8*\elmtsize+\offset, -0.5*\elmtsize) {Input\\Batch};
        \node[align=center] () at(11*\elmtsize+2*\offset, -0.5*\elmtsize) {Bias};

    \end{scriptsize}
\end{tikzpicture}
    \caption{Illustration of the linear operations for batch processing. On the left is the standard batch processing operation with a linear product of a $\mathbb{R}^{6 \times 5}$ weight matrix, an $\mathbb{R}^{5 \times 3}$ input with a batch size of 3, and a $\mathbb{R}^{6 \times 1}$ bias vector duplicated for each batch. On the right is the modified approach. The three $\mathbb{R}^{6 \times 5}$ weight matrices corresponding to a batch size of 3 form a $\mathbb{R}^{6 \times 5 \times 3}$ tensor, the three bias vectors corresponding to a batch size of 3 form a $\mathbb{R}^{6 \times 1 \times 3}$ tensor, and the input batch matrix is rearranged into a $\mathbb{R}^{5 \times 1 \times 3}$ tensor. The matrix operations are standard matrix/vector operations over the first and second dimensions of the tensors. (For illustrative purposes, the batch dimension is last, but in practice it is more appropriate to have it first.)}
    \label{fig:batchProcessing}
\end{figure}

The standard approach to using minibatches in stochastic gradient descent can be used in the unconditioned posterior model. However, in the conditional model, if $\acute{\x}$ is provided in batch form, the output of the posterior model will be a batch of primary model parameters $\thetabf$: a set of primary parameters associated with each $\acute{\x}$ in the batch. As the batched inputs are passed through the primary model, each element in the batch be associated with the corresponding element in parameter batch. This is easily achieved via matrix multiplication, however, the tensors containing the variables should be carefully arranged. An illustration comparing standard batch processing and the modified batch processing is provided in \figurename{~\ref{fig:batchProcessing}}. Note that the modified batch approach is equivalent to stochastic gradient descent with a single sample. The result is that the only limit on batch size is memory capacity.
%

\section{Posterior Distribution Model Configurations}
\label{sec:posteriorDistributionModelConfigurations}

\begin{figure}[!b]
    \centering
    \includegraphics{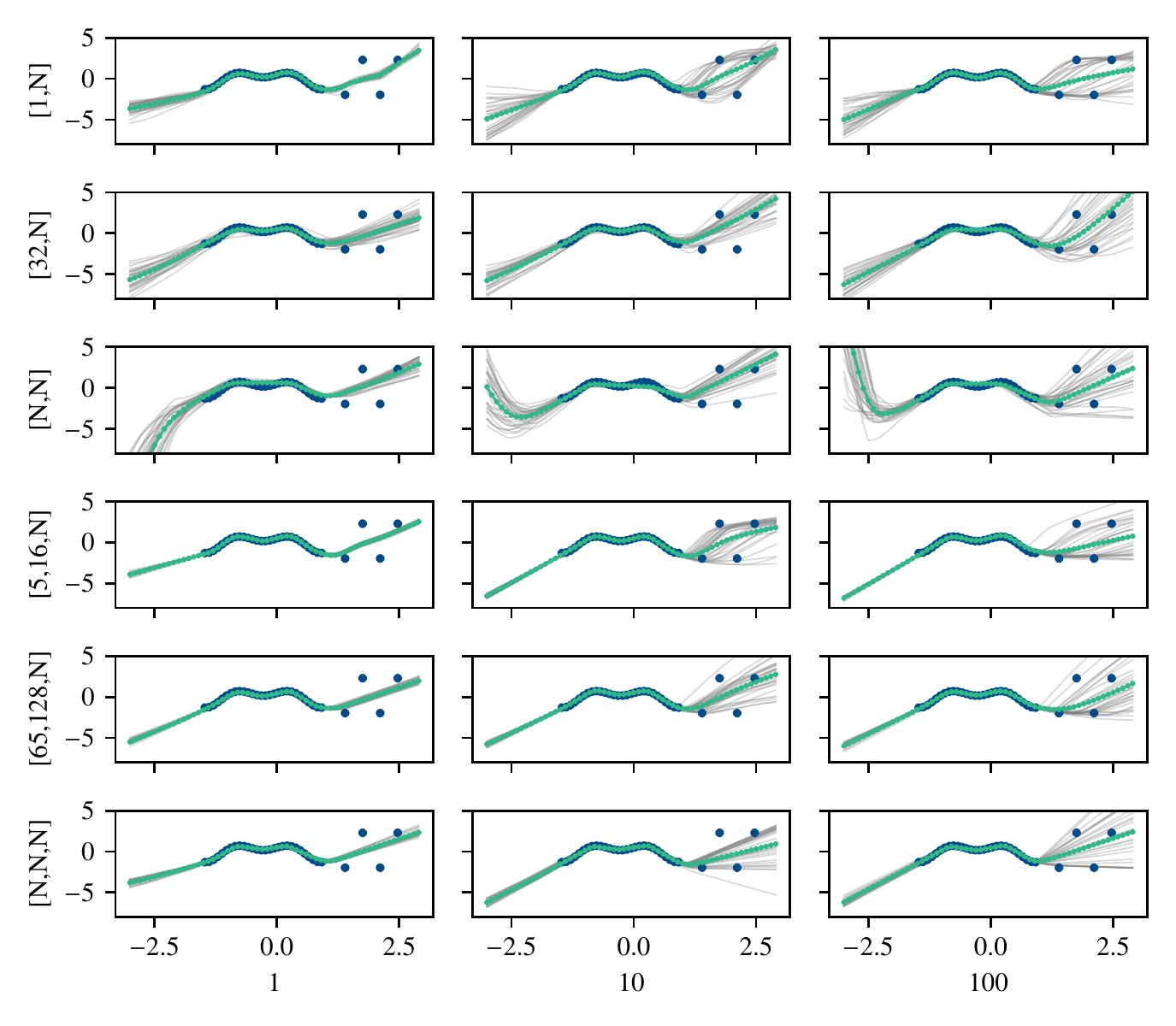}
    \caption{Unconditioned posterior model. Columns are plots for 1, 10, and 100 MC samples used during training. Rows plot results with variations in the posterior model architecture, where the brackets denote the number of units in the layers of the network. All models have $N$ outputs. Blue markers are the targets, the grey curves are 30 MC samples from the posterior predictive distribution, and the green curve is the mean of these samples. Note that these figures plot standardised data.}
    \label{fig:post_test_results_MlpRegressor_UnconditionedLinear}
\end{figure}

\begin{figure}[!t]
    \centering
    \includegraphics{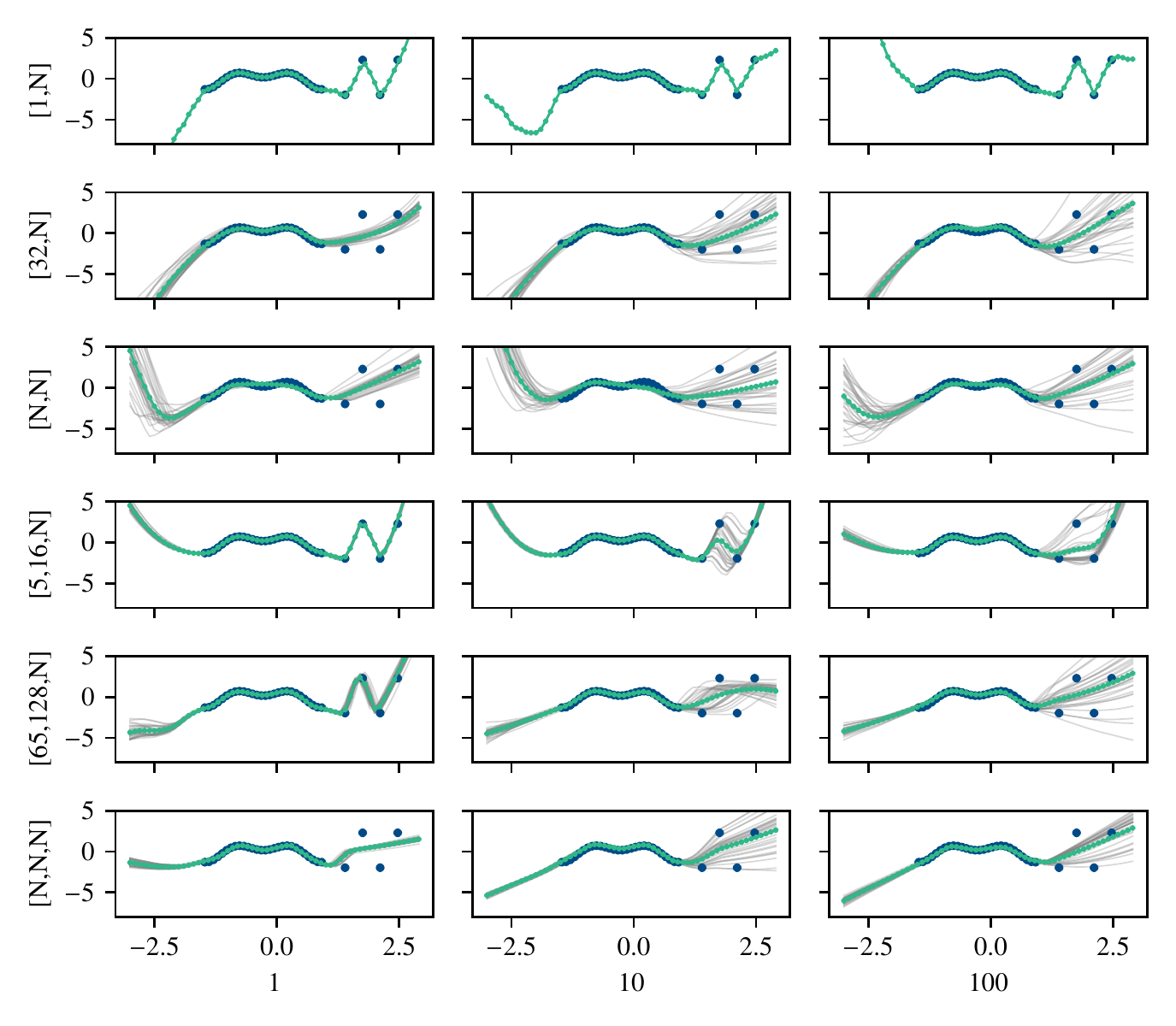}
    \caption{Conditional posterior model. Columns are plots for 1, 10, and 100 MC samples. Rows plot results with 1 input ([1]), 32 inputs ([32]), 5 inputs with 16 hidden neurons ([5,16]), $N$ inputs ([NN]), and $N$ inputs with $N$ hidden layers ([NNN]). All models have $N$ outputs. Blue markers are the targets, the grey curves are 30 MC samples from the posterior predictive distribution, and the green curve is the mean of these samples. Note that these figures plot standardised data.}
    \label{fig:post_test_results_MlpRegressor_ConditionalLinear}
\end{figure}

We consider varying the architecture and number of MC samples used in training for the posterior model. Results are plotted in \figurename{} \ref{fig:post_test_results_MlpRegressor_UnconditionedLinear} and \figurename{} \ref{fig:post_test_results_MlpRegressor_ConditionalLinear}. The columns specify the number of Monte Carlo samples $L$ used when training the model. The rows specify a configuration of the posterior model using a list containing two or three integers. In the case of two integers, the posterior model contains an input layer, an output layer, and no hidden layer. As such, the posterior model provides a linear translation of its inputs $\z$. In the case where three integers are provided in the list, a MLP with one hidden layer is used, where the number of units in the hidden layer are provided by the second integer in the list. When a hidden layer is used, the ReLU activation function applies.

Considering the number of MC samples, a single MC sample does not seem to provide sufficient capacity to model the posterior distribution. However, for this dataset, there is no significant difference between 10 and 100 MC samples, suggesting that 10 samples are sufficient. In general, this is confirmed by \citet{bishop2006pattern}, who suggests that MC estimate of an integral only requires a few samples if the samples are i.i.d. In the case of the posterior models, $\z$ are indeed i.i.d. randomly generated samples.

The first three rows in the plots use a linear transfer function where no hidden layer in the posterior model is used. Especially for the unconditioned models, the out-of-distribution variation to the left of the data is larger for the linear models than for the MLP models. The higher complexity seems to encourage more certainty around that region. By conditioning on the inputs $\acute{\x}$, it does seem to improve the uncertainty. Furthermore, owing to the conditioning on $\acute{\x}$, the conditional models generally produce more complex curves around the high variance samples.

When the posterior model maps from a lower dimension to a higher dimension, the posterior distribution will lie on a union of manifolds within the domain of $\thetabf$ \cite{arjovsky2017towards}.  With the highest model complexity $[N,N,N]$, the posterior is able to span the domain of $\thetabf$. The similarity of the results from models with a small number of dimensions (e.g. the $[1]$ and $[5,16,N]$ models) with the $[N,N,N]$ posterior model suggest that these models are still able to provide reasonable uncertainty estimates. A smaller posterior model can significantly reduce the overall computational complexity.

%
%
%
    

\section{Prior Inclusion Results}
\label{sec:priorInclusionDemonstrationAppendix}

We demonstrate including a prior by adding the sum-of-squares regulariser in the loss function presented in (\ref{eq:post_pred_loss}). If the likelihood function is chosen as the sum of squared errors the loss function is given by
\begin{align}
    \label{eq:regularizer}
    \mathcal{L}
    &= - \frac{1}{L} \sum_{l=1}^L \bigg( \underbrace{-\sum_{i=1}^N (y_i - y_i')^2}_{\text{likelihood}} - \underbrace{ \vphantom{\sum_{i=1}^N} \lambda \left( \thetabf^{(l)} \right)^T \thetabf^{(l)}}_{\text{regulariser}} \bigg)
\end{align}
Note the resemblance of contents within the parentheses to the maximum a posterior (MAP) estimate where $\lambda ( \thetabf^{(l)} )^T \thetabf^{(l)}$ relates to a zero mean isotropic Gaussian with precision $2\lambda$. The key difference is that this regularized loss function is computed within a MC sum over parameters sampled from the posterior distribution.

\begin{figure}[!tb]
    \centering
    \includegraphics{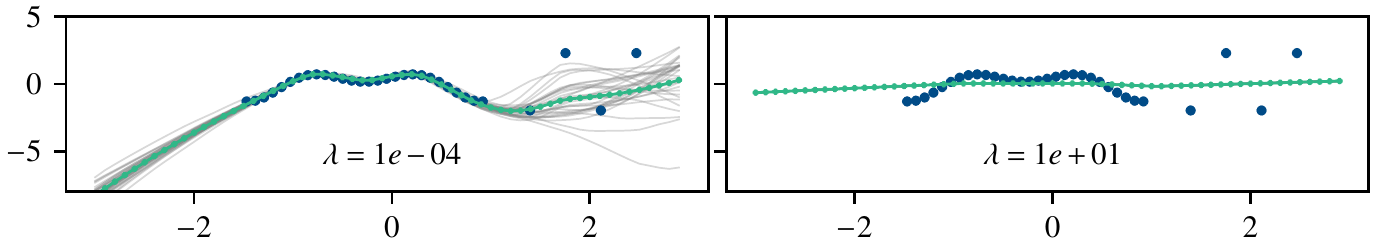}
    \caption{Results with a regulariser introduced for prior information. Blue markers are the targets, the grey curves are 30 MC samples from the posterior predictive distribution, and the green curve is the mean of these samples. Note that these figures plot standardised data.}
    \label{fig:resultsPriorAppendix}
\end{figure}

The results are presented in \figurename{~\ref{fig:resultsPriorAppendix}}. Comparing these results to the results with no regularisation presented in \figurename{~\ref{fig:results}}, the left panel shows slight under-fitting of the data owing to the regularization. The variance increases significantly around the high variance samples and the out-of-distribution regions. Note the similarity of this result to the result presented for the linear model in \figurename{~\ref{fig:resultsLr}}. This illustrates that the regularization is penalising the complexity of the primary model as desired. When the regularisation constant $\lambda$ is increased to an extreme value, the model produces a horizontal line for all MC samples as illustrated in the right panel. Here the model has fitted the data with an offset, which corresponds to the mean of the targets $\y$ (which is zero). With such high regularisation, the model be forced into a restricted region of $\thetabf$ resulting in a degenerate Dirac distribution for the posterior predictive distribution as described in Section \ref{sec:prior} and Appendix \ref{sec:degeneratePosterior}. 
%

\end{document}